%% file: main.tex
\documentclass[10pt,twocolumn,letterpaper]{article}

\usepackage[pagenumbers]{wacv} 

\usepackage{pgfplots}


\definecolor{wacvblue}{rgb}{0.21,0.49,0.74}
\usepackage[pagebackref,breaklinks,colorlinks,allcolors=wacvblue]{hyperref}

\def\wacvPaperID{81} 
\def\confName{WACV}
\def\confYear{2026}

\title{Salience-SGG: Enhancing Unbiased Scene Graph Generation with Iterative Salience
Estimation}

\author{
Runfeng Qu$^{1,5}$ \quad
Ole Hall$^{1,5}$ \quad
Pia K Bideau$^{3}$ \quad
Julie Ouerfelli-Ethier$^{2,5}$ \quad\\
Martin Rolfs$^{2,5}$ \quad
Klaus Obermayer$^{1,4,5}$ \quad
Olaf Hellwich$^{1,5}$ \\
$^{1}$Technische Universität Berlin, Germany
$^{2}$Humboldt Universität zu Berlin, Germany \\
$^{3}$Univ. Grenoble Alpes, Inria, CNRS, Grenoble INP, LJK, France \\
$^{4}$Bernstein Center for Computational Neuroscience, Germany\\
$^{5}$Science of Intelligence Research Cluster of Excellence, Germany\\
{\tt\small runfeng.qu@campus.tu-berlin.de}
}

\begin{document}
\maketitle
\input{sec/0_abstract}    
\input{sec/1_intro}
\input{sec/2_relatedwork}
\input{sec/3_Method}
\input{sec/4_experiments}
\input{sec/5_conclusion.tex}
    

\input{main.bbl}
\end{document}

%% file: sec/0_abstract.tex
\begin{abstract}
Scene Graph Generation (SGG) suffers from a long-tailed distribution, where a few predicate classes dominate while many others are underrepresented, leading to biased models that underperform on rare relations. Unbiased-SGG methods address this by implementing debiasing strategies, but often at the cost of spatial understanding—resulting in over-reliance on semantic priors. We introduce Salience-SGG, a novel framework featuring an Iterative Salience Decoder (ISD) that emphasizes triplets with salient spatial structures. To support this, we propose semantic-agnostic salience labels guiding ISD. Evaluations on Visual Genome, Open Images V6, and GQA-200 show that Salience-SGG achieves state-of-the-art performance and improves existing Unbiased-SGG methods in their spatial understanding as demonstrated by the Pairwise Localization Average Precision. Code is available at: \hyperlink{Salience-SGG}{https://github.com/runfeng-q/Salience-SGG}.
\end{abstract}

%% file: sec/1_intro.tex
\section{Introduction}
\label{sec:intro}
For any given image, scene graph generation (SGG) models aim to detect a set of triplets in the form of $\{subject, predicate, object\}$. These triplets can be transformed into graph structures called scene graphs, where nodes indicate subjects and objects, while edges represent predicates. 
The primary challenge in SGG tasks is the long-tailed distribution problem. Unbalanced predicate samples result in predictions of SGG models dominated by a few frequent predicates (\eg wearing, has), while undermining the rare predicates (\eg lying on, flying in). Recent SGG studies focus on Unbiased Scene Graph Generation (Unbiased-SGG), which aims to create fine-grained scene graphs with more informative predicates by incorporating various debiasing methods such as re-weighting \cite{li2022ppdl,chen2022resistance}, re-sampling \cite{li2021bipartite, dong2022stacked}, and pseudo-labeling \cite{zhang2022fine, kim2024adaptive, goel2022not}.
\begin{figure}[t]
    \centering
    \includegraphics[width=\linewidth]{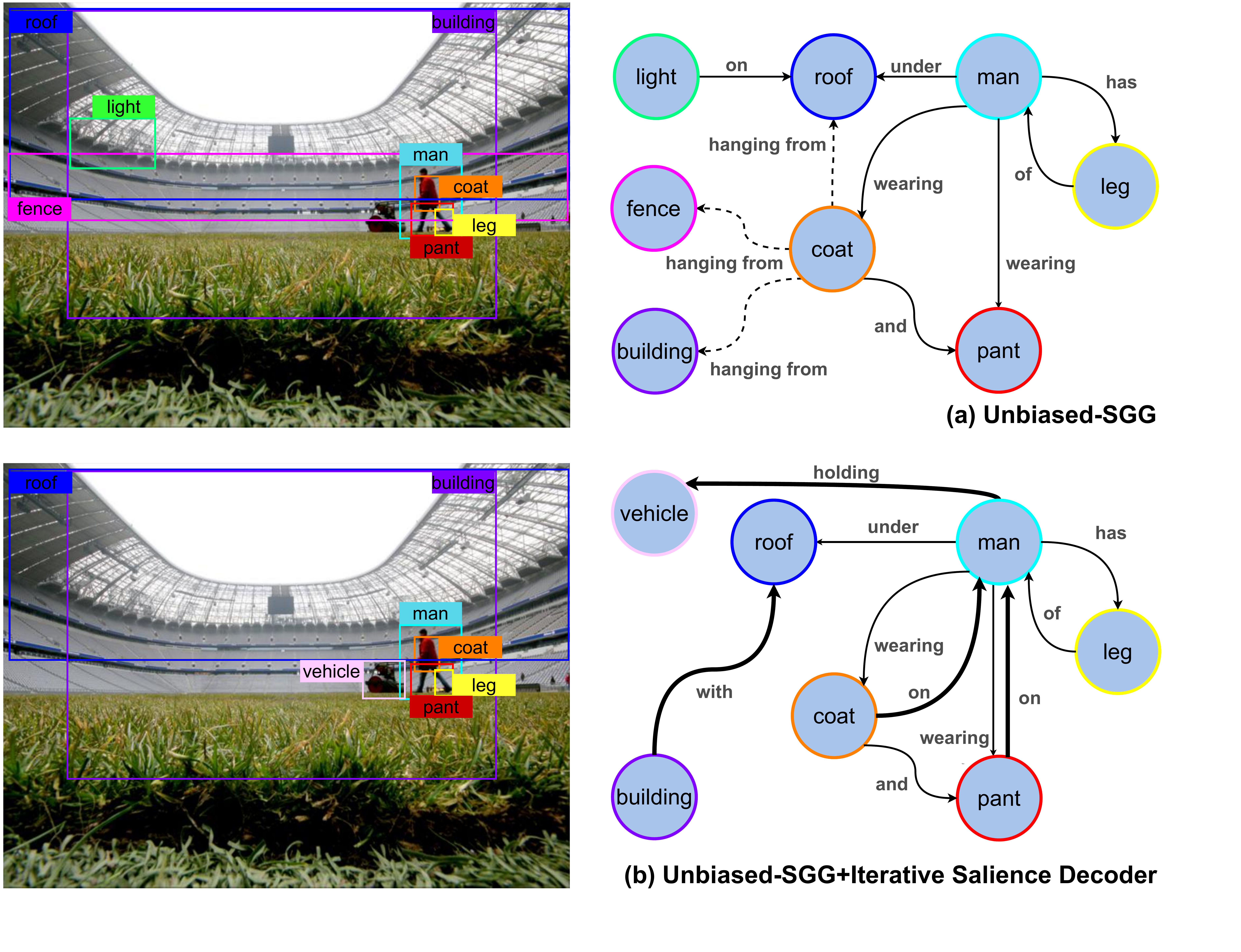}
\caption{Salience-SGG. (a): IETrans~\cite{zhang2022fine} with standard debiasing, showing over-reliance on the semantic information, \ie \textit{coat} and \textit{hanging from} (dashed lines). (b): Our salience-enhanced model favors spatially coherent triplets (bold).}
    \label{fig: Intro_image}
\end{figure}

Despite the progress made in Unbiased-SGG studies, prevalent models do not demonstrate strong robustness to debiasing strategies. Improving performance on rare predicates often substantially reduces performance on frequent ones \cite{li2021bipartite}. This phenomenon can be attributed to the model's preference for rare predicates, encouraged by debiasing strategies. For triplets such as $\{person, on, beach\}$, Unbiased-SGG models may incorrectly favor rare predicates (e.g., standing on). This example illustrates the trade-off that undermines performance on common predicates. However, we identify that debiasing strategies disrupt spatial understanding of entity pairs. Specifically, spatially loosely associated entity pairs with rare predicates are detected with high confidences, as illustrated in \Cref{fig: Intro_image} (a). These spatially incoherent pairs, in turn, suppress other entity pairs with more salient spatial structures to be recalled, thereby further decreasing overall performance. The reason for this issue is that the prevailing debiasing methods rely solely on the frequency statistics of predicate and entity semantics. Applying these methods to detection tasks may result in a model that relies significantly on high-level semantic information of entity pairs alone, while disregarding low-level spatial information for global detection. As a consequence, Unbiased-SGG models may not capture the triplets in salient spatial structures. We refer to this issue as \textbf{salience insensitivity}, given the lesser sensitivity to the spatial information. To analyze salience insensitivity systematically, we introduce a measurement that evaluates the ability of SGG models to capture salient triplet structures (see \Cref{Spatial Understanding Analysis}).

To address the problem of salience insensitivity, we propose a novel framework that integrates a triplet salience estimation process into the Unbiased-SGG process. The salience estimation is achieved with the introduction of triplet salience labels and an Iterative Salience Decoder (ISD). The triplet salience labels are expressed by a mask in the form of $\{0,1\}$, which indicates whether any two detected entities form a salient triplet that spatially matches the ground-truth triplets. Our triplet salience labels and ISD stand in contrast to existing approaches \cite{wang2024multi, im2024egtr,jung2023devil}, which directly address the corresponding annotated triplets involving semantic content to create top-down binary labels. We generate bottom-up labels that consider only the low-level spatial information of the ground-truth triplets. The rich spatial features contained within these salience labels provide ISD with significant guidance for emphasizing the spatial configuration and globally exploring the triplet candidates in most important spatial structures within scenes, as shown in \Cref{fig: Intro_image} (b). Furthermore, such semantic-agnostic labels prevent the salience estimation from becoming a prediction of whether each triplet is positive. 
ISD formulates the triplet salience estimation as a salience message-passing process. To effectively capture the important spatial structures in scenes, we propose two types of enhanced attention layers to guide the message-passing. In summary, our contributions are as follows: 
\begin{itemize}
  \item We propose a novel framework for Unbiased-SGG that incorporates salience estimation to directly address the \textbf{salience insensitivity} caused by debiasing strategies.
  \item We introduce \textbf{bottom-up salience labels}, and \textbf{enhanced attention layers} to facilitate the exploration of the salient spatial structures of the visual scene by an iterative salience decoder.
  \item We provide extensive experimental results on three challenging dataset demonstrating that our proposed framework improves the performance on the Unbiased-SGG task by enhancing salience sensitivity.
\end{itemize}

%% file: sec/2_relatedwork.tex
\section{Related Work}
\label{sec:Related Works}
\textbf{Scene Graph Generation.} Early work on SGG follows the two-stage framework which contains an object detector such as Faster R-CNN \cite{ren2016faster} to propose entities. A predicate predictor is performed on the entity pairs to predict the predicates between them \cite{lu2016visual, zhang2017visual}. Subsequently, hierarchical structures are proposed \cite{dai2017detecting, xu2017scene,yin2018zoom} to jointly update entity pairs and predicates. \cite{zellers2018neural, woo2018linknet} employ Long Short-Term Memory (LSTM) \cite{hochreiter1997long} and Attention Mechanism \cite{vaswani2017attention} to aggregate global context for entity updating and predicate prediction. Inspired by the Detection Transformer (DETR) \cite{carion2020end}, several one-stage SGG models \cite{khandelwal2022iterative, teng2022structured, Dong_2021_ICCV} are developed achieving SGG by maintaining a set of learnable queries. 

Recall (R@K) is the standard metric for evaluating SGG models, but it fails to reflect performance on rare predicates due to the long-tailed distribution. To address this, mean recall (mR@K) ~\cite{tang2019learning}  equally weigh all predicate classes to better assess generalization.
To achieve high mR@K, \cite{tang2020unbiased} propose an Unbiased-SGG model with causal effects. BGNN \cite{li2021bipartite} re-balances the dataset using a bi-level re-sampling strategy. Various studies \cite{chen2022resistance, khandelwal2022iterative, sudhakaran2023vision} utilize the re-weighting losses to adjust the decision boundary for each individual predicate. Finally, inspired by knowledge distillation, many recent studies \cite{yoon2025ra, zang2024refine} focus on creating pseudo-labels based on predictions of a pre-trained SGG model to enrich annotations.  

Current Unbiased-SGG models focus on optimizing mR@K, often overlooking the negative impact of debiasing strategies on R@K. To balance both, IETrans \cite{zhang2022fine} propose F@K as an overall performance, which is the harmonic mean of R@K and mR@K. Mg-RMPN \cite{wang2024multi} discover that using a binary classification module to select likely triplets may greatly benefit overall performance.
However, the binary classification module is trained using binary labels that are consistent with the annotated triplets. The high correlation between these labels and triplet semantic information may not sufficiently support the model exploring salient spatial structure.
Our salience estimation module is trained using labels derived solely from the spatial information of ground-truth graphs. These semantic-agnostic labels guide the module to focus purely on spatial patterns.

\textbf{Salience-Guided Object Detector.} 
Salience estimation is commonly used to improve object detection tasks. Salience-DETR\cite{hou2024salience} and Focus-DETR \cite{zheng2023less} integrate feature-level salience modules into DETR to reduce the computational burden. Yolo-sg \cite{han2022yolo} creates pixel-level salience labels to help the detector focus on important regions of images.  Inspired by these studies, we propose Salience-SGG, which encourages exploration of important spatial structures by leveraging triplet-level salience labels.

%% file: sec/3_method.tex
\section{Method}
\label{Method}
\begin{figure*}[t]
    \centering
    \includegraphics[height=0.28\textheight]{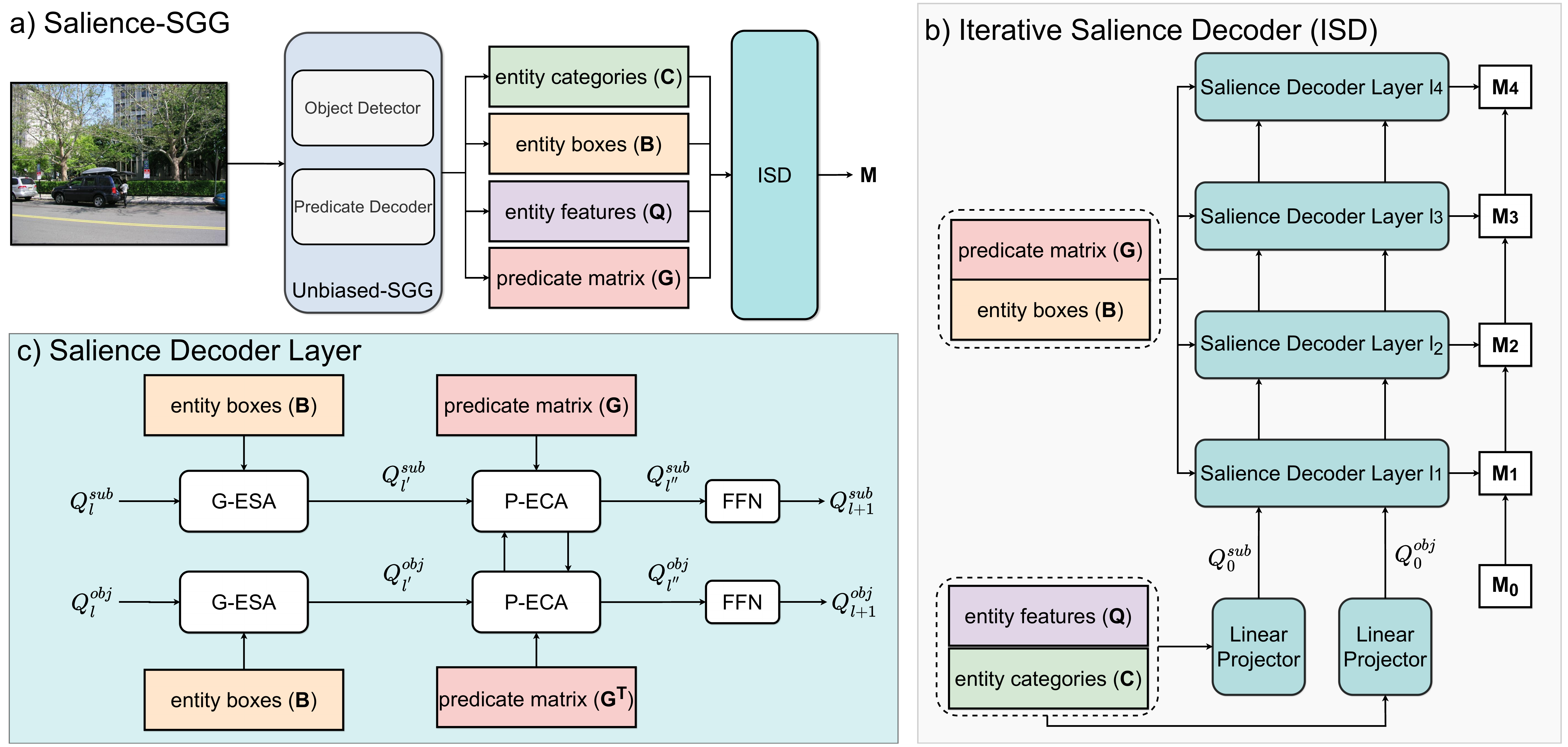}
    \caption{(a) Overview of our full network architecture. Our framework consists of an Unbiased-SGG module and a model-agnostic iterative salience decoder (ISD) module. (b) Subject salience features and object salience features are updated through multiple salience decoder layers in ISD. (c) Details of the message-passing process of the salience decoder layer. G-ESA and P-ECA correspond to Geometry Enhanced Self-Attention and Predicate Enhanced Cross-Attention, respectively.
    }
    \label{fig: architechture}
\end{figure*}
SGG tasks aim to detect a scene graph represented as $ {\mathcal{G}}=\{{\mathcal{\hat{V}}}, {\mathcal{\hat{E}}} \}$, where ${\mathcal{\hat{V}}}$ denotes the set of entities and ${\mathcal{\hat{E}}}$ is the set of triplets. Each entity $\hat{v}_{x} \in {\mathcal{\hat{V}}}$ contains information about its category and location, \ie$\hat{v}_{x}=\{\hat{b}_{x}, \hat{c}_{x}\}$. In general, each triplet $\hat{e}_{k} \in {\mathcal{\hat{E}}}$ is represented in the form of $\{\hat{s}_{k},\hat{p}_{k}, \hat{o}_{k}\}$, where $\hat{s}_{k}, \hat{o}_{k}$ refer to the 
entity indices in ${\mathcal{\hat{V}}}$ of the corresponding subject and object, while $\hat{p}_{k}$ indicates the predicate category connecting the entity pair. To address salience insensitivity, we reformulate the triplet as $\{\hat{s}_{k},\hat{p}_{k}, \hat{o}_{k}, \hat{m}_{k}\}$, where $\hat{m}_{k}$ is a binary label indicating triplet salience.

The overall framework of our approach consists of an Unbiased-SGG module and a salience decoder module, \ie Iterative Salience Decoder (ISD), as depicted in \Cref{fig: architechture} (a). To investigate the impact of ISD on various debiasing strategies, we provide both one-stage and two-stage methods. The implementation of the two-stage method is achieved through the integration of ISD into the extant two-stage Unbiased-SGG models (see \Cref{Complementary Comparisons}). Unless specified otherwise, the following sections refer to our one-stage method \textbf{Salience-SGG}. In the Unbiased-SGG part, a deformable DETR \cite{zhu2020deformable} is employed as an object detector to propose a set of entities ${\mathcal{V}}=\{v_{i}\}_{i=1}^{N_{e}}$, where $N_{e}$ is the number of detected entities. Each detected entity $v_{i}$ contains its bounding box location and category distribution, denoted as ${B}=\{ b_{i}\}^{N_{e}}_{i=1} \in\mathbb{R}^{N_{e} \times 4}$, ${C}=\{ c_{i}\}^{N_{e}}_{i=1} \in\mathbb{R}^{N_{e} \times N_{c}}$, respectively, where $N_{c}$ is the total number of entity categories. In addition, we extract the queries from the last decoder layer of the deformable DETR as entity features ${Q}=\{ q_{i}\}^{N_{e}}_{i=1} \in \mathbb{R}^{N_{e} \times d}$, where $d$ is the dimension of these features. The object detector is followed by a predicate decoder to predict the predicate connecting each detected entity pair, denoted as ${G} \in {\mathbb{R}^{N_{e} \times N_{e} \times N_{p}}}$ (see \Cref{predicate classifier}). Following previous work  \cite{im2024egtr, wang2024multi}, the entity labels and predicates labels for the detected entities are obtained by applying Hungarian Matching \cite{kuhn1955hungarian}. Entity labels are denoted as $V^{'}=\{v^{'}_{i}\}_{i=1}^{N_{e}}$, where $v_{i}^{'}=\{b^{'}_{i}, c^{'}_{i}\}$, while the predicate labels are indicated as ${G^{'}} \in {\mathbb{R}^{N_{e} \times N_{e} \times N_{p}}}$, where $N_{p}$ is the category number of predicates. Leveraging the ground-truth graph ${\mathcal{G}}$, we construct the triplet salience labels ${M^{'}} \in {\mathbb{R}^{N_{e} \times N_{e}}}$ as described in \Cref{Triplet Salience Label}. Ultimately, ${G}, {C}, B, Q$ are 
sent to the ISD to estimate the triplet salience intensity ${M} \in {\mathbb{R}^{N_{e} \times N_{e}}}$ and approximate ${{M^{'}}}$ (for details see \Cref{Iterative Salience Decoder}).

\subsection{Predicate Decoder}
\label{predicate classifier}
Our Salience-SGG uses a lightweight predicate decoder to reason about predicates directly. Given the entity information, the predicate features $R=\{r_{ij}\} \in \mathbb{R}^{N_{e} \times N_{e} \times D}$ are created as written in \Cref{triplet features}. Subsequently, $R$ is transmitted to a MLP classifier to obtain $G$ without any further refinement process.
\begin{equation}
\label{triplet features}
    r_{ij}=[b_{i},\text{GloVe}_{i}, q_{i}, b_{j},\text{GloVe}_{j}, q_{j}]
\end{equation}
where $[\cdot, \cdot]$ denotes the operation of concatenation, and Glove indicates the glove embedding \cite{pennington2014glove} of the predicted entity category from the object detector.

\subsection{Triplet Salience Label}
\label{Triplet Salience Label}
The triplet salience label is created to guide our ISD to explore the significant spatial structures related to the annotated graphs, thereby maintaining salience sensitivity during the Unbiased-SGG process. In order to preserve the geometric relationships between subjects and objects, we use the entity-level locations of the annotated triplets as a criterion to assign salience labels. A given triplet candidate is considered salient when both the subject and object boxes sufficiently overlap with the subject and object of any one of the ground-truth triplets at the same time. Particularly, the locations of subjects and objects in the ground-truth triplets set are denoted by $\{b^{'}_{s_{k}}, b^{'}_{o_{k}}\}_{k=1}^{K}$, where $K$ is the number of annotated triplets. $\{ b_{i}\}^{N_{e}}_{i=1}$ refers to the bounding boxes of the detected entities. A subject salience matrix $\hat{M}^{sub} \in \mathbb{R}^{N_{e} \times K}$ is constructed by computing the pairwise Intersection over Unions (IoUs) between $\{ b_{i}\}$ and $\{b^{'}_{s_{k}}\}$. Similarly, an object salience matrix $\hat{M}^{obj} \in \mathbb{R}^{N_{e} \times K}$ reflects the pairwise IoUs between $\{ b_{i}\}$ and $\{b^{'}_{o_{k}}\}$. The salience labels of detected entity pairs are determined as follows:
\begin{equation}
\centering
M^{'}_{ij} = 
\left\{
\begin{array}{ll}
1, & \exists{a},  \hat{M}^{sub}_{ia}>=\mathcal{T} \And \hat{M}^{obj}_{ja}>=\mathcal{T} \\
0,   & \text{otherwise}
\end{array}
\right.
\end{equation}
Due to the omission of semantic information regarding entity pairs during salience label assignment, a significant number of triplet candidates with low semantic information or incoherent semantics are considered salient. This kind of label forces ISD to rely heavily on spatial information when making estimations.
\subsection{Iterative Salience Decoder}
\label{Iterative Salience Decoder}
The structure of ISD is illustrated in \Cref{fig: architechture} (b). Using the detected entities $Q$, $B$, and $C$ as well as the predicate matrix $G$, ISD estimates the salience of each entity pair ${M \in {\mathbb{R}^{N_{e} \times N_{e}}}}$. Because each image contains $N_{e} \times N_{e}$ entity pairs, globally estimating the salience by maintaining $N^{2}_{e}$ triplet salience features brings significant computational cost. In this study, we introduce subject salience queries $Q^{sub} \in \mathbb{R}^{N_{e} \times d}$ and object salience queries $Q^{obj} \in \mathbb{R}^{N_{e} \times d}$ to preserve subject and object salience information. $Q^{sub}$ and $Q^{obj}$ are updated through a message-passing process which is achieved by multiple salience decoder layers. Each layer of our iterative salience decoder contains two branches for updating $Q^{sub}$ and $Q^{obj}$ respectively (see \Cref{fig: architechture} (c)). Each branch consists of a Geometry Enhanced Self-Attention (G-ESA), a Predicate Enhanced Cross-Attention (P-ECA) and a Feed Forward Network (FFN). The final triplet salience ${M}$ is estimated by fusing $Q^{sub}$ and $Q^{obj}$.

\textbf{Subject and Object Salience Queries Initialization. } We denote the updated subject and object salience queries after $l$-th salience decoder layer as $Q^{sub}_{l}$ and $Q^{obj}_{l}$ respectively. Since ISD is a module to model the salience message-passing process, the initial salience queries $Q^{sub}_{0}$ and $Q^{obj}_{0}$ reflect the salience information of individual entities as determined by the object detector. Instead of creating additional learnable vectors, we design a set of entity salience features $Q^{ent} \in \mathbb{R}^{N_{e} \times d}$ with the entity category distributions $C$ and features $Q$ obtained from object detection as follows:
\begin{equation}
\label{salience queries}
    Q^{ent}=Q+\text{proj}(C)
\end{equation}
where $\text{proj}$ indicates a linear projector to map the category distribution to the same dimension as the entity features. Subsequently, two linear projectors are applied to $Q^{ent}$ to return $Q^{sub}_{0}$ and $Q^{obj}_{0}$.

\textbf{Geometry Enhanced Self-Attention.} G-ESAs are adopted to conduct intra-subject and intra-object salience message-passing. Before introducing our G-ESA, let us revisit the process of the regular self-attention process. Given the subject salience queries $Q^{sub}_{l}$, the regular self-attention first computes an attention matrix $A_{self} \in \mathbb{R}^{N_{e}\times N_{e} \times H}$, where H is the head number of self-attention:
\begin{equation}
\label{AM}
    A_{self}=\frac{\text{Que}(Q^{sub}_{l}) \cdot \text{Key}(Q^{sub}_{l})^{T}}{\sqrt{d}}
\end{equation}
The queries are updated according to $A_{self}$ as follows:
\begin{equation}
\label{SA}
    Q^{sub}_{l^{'}}=Q^{sub}_{l}+\text{SoftMax}(A_{self}) \cdot \text{Val}(Q^{sub}_{l})
\end{equation}
The regular self-attention layers control the information flow solely based on the subspace similarity among queries as shown in \Cref{AM}. According to the definition of our triplet salience described in \Cref{Triplet Salience Label}, subjects or objects exhibiting high overlap areas should be characterized by similar salience. Inspired by \cite{hao2023relation, hou2024relation}, we introduce G-ESA to reinforce the information interaction between the queries determined by their spatial overlapping.
Specifically, we calculate the IoUs between every entity pairs forming $\mathcal{U} \in \mathbb{R}^{N_{e} \times N_{e}}$. Subsequently, $\mathcal{U}$ is projected to equal the channel to the head number $H$ of the self-attention layer. Ultimately, the projected $\mathcal{U}$ is incorporated into the attention matrix $AM_{self}$ in order to manipulate the information flow as below:
\begin{equation}
\label{new_AM}
  A_{self}=\text{Relu}(\text{MLP}(\mathcal{U}))+\frac{\text{Que}(Q^{sub}_{l}) \cdot \text{Key}(Q^{sub}_{l})^{T}}{\sqrt{d}}
\end{equation}
The same operation is used for intra-object salience message-passing to obtain $Q^{obj}_{l^{'}}$. 
Note that, to ensure training stability, a ReLU activation function is applied after the MLP output layer to suppress negative values.

\textbf{Predicate Enhanced Cross-Attention.} we introduce P-ECA to conduct inter-subject-object salience message-passing. Similarly to G-ESA, the interactions between subjects and objects are enhanced in accordance with the predicted predicates $G$. For instance, in the subject branch, subject salience queries are updated as follows:
\begin{equation}
  A_{cross}=\text{Relu}(\text{MLP}(G))+\frac{\text{Que}(Q^{sub}_{l^{'}}) \cdot \text{Key}(Q^{obj}_{l^{'}})^{T}}{\sqrt{d}}
\end{equation}
\begin{equation}
\label{cross} Q^{sub}_{l^{''}}=Q^{sub}_{l^{'}}+\text{SoftMax}(A_{cross}) \cdot \text{Val}(Q^{obj}_{l^{'}})
\end{equation}
The operation is identical for updating object salience queries, with the exception that $G$ is first transposed to $G^{T}$. After P-ECAs, FFNs performed on $Q^{sub}_{l^{''}}$ and $Q^{obj}_{l^{''}}$ lead to the two following outputs: $Q^{sub}_{l+1}$ and $Q^{obj}_{l+1}$.

\textbf{Iterative Triplet Salience Refinement.} As demonstrated by our empirical results, an iterative refinement process yields superior results in comparison to directly considering the predictions after each salience decoder layer as the triplet salience. To reduce computational costs and in line with the salience message-passing process in the salience decoder layer, linear attention is utilized to fuse the subject and object salience queries. These fused queries are added to the triplet salience predicted from the previous layer to form the refined triplet salience as follows:
\begin{equation}
\label{salience intensity}
    M_{l+1}=\text{sigmoid}(\text{inverse\_sigmoid}(M_{l})+ \frac{Q^{sub}_{l+1} \cdot (Q^{obj}_{l+1})^{T}}{\sqrt{d}})
\end{equation}
where $\text{sigmoid}$ refers to the sigmoid activation function. For $l=0$, the initial salience matrix $M_0$ is a zero matrix.

\subsection{Training and Inference}
For the entire training process of our Salience-SGG, the overall loss function $\mathcal{L}$ is expressed as follows: 
\begin{equation}
\begin{aligned}
&\mathcal{L}=\mathcal{L}_{ent}+\mathcal{L}_{salience}+\mathcal{L}_{pre} \\
  &\mathcal{L}_{ent} =\mathcal{L}^{cls}_{ent}+\mathcal{L}^{box}_{ent}+\mathcal{L}^{GIoU}_{ent}
\end{aligned}
\end{equation}
where $\mathcal{L}_{ent}$ is the entity loss, which is identical to the loss function of the deformable DETR \cite{zhu2020deformable}. We compute focal loss \cite{lin2017focal} between $M$ and $M^{'}$ as the salience loss $\mathcal{L}_{salience}$. Following \cite{wang2024multi}, we employ re-weighting loss called seesaw loss \cite{wang2021seesaw} as the predicate loss $\mathcal{L}_{pre}$ in our Salience-SGG to solve the long-tailed distribution problem.

For inference, we follow \cite{im2024egtr}, except that triplet candidates are sorted by involving our salience scores $M$.

%% file: sec/4_experiments.tex
\section{Experiments}
\label{experiments}
\subsection{Experimental Setup}
This section provides descriptions of datasets, evaluation metrics, and implementation details.

\textbf{Datasets.} We conduct experiments on three datasets: 
\newline \textit{Visual Genome (VG)} \cite{krishna2017visual} contains 57,723 training, 5,000 validation, and 26,446 test images. It comprises of 150 entity and 50 predicate categories.
\newline \textit{Open Image V6 (OIv6)} \cite{kuznetsova2020open} contains 126,368 training, 1,813 validation, and 5,322 test images, totaling 133,503 images across 601 entity and 30 predicate categories.
\newline \textit{GQA-200} \cite{hudson2019gqa} consists of 57,623 training, 5,000 validation, and 8,208 test images. After processing, GQA-200 is composed of 200 entity and 100 predicate categories.
\newline
For all three datasets, we follow the pre-processing in \cite{wang2024multi}.

\textbf{Evaluation Metrics.}
For each dataset, we present the performance on the scene graph detection (SGDet) task. In the case of the VG and GQA-200 datasets, we follow previous work \cite{zhang2022fine} reporting Recall@K (R@K),  mean Recall@K (mR@K), and F@K to measure the overall performance. For the OIv6 dataset, we include micro-Recall@50 (micro-R@50), weighted mean AP of triplets (wmAPrel), weighted mean AP of phrase (wmAPphr), and the final score. The final score is calculated by score = 0.2 × Recall@50 + 0.4 × wmAPrel + 0.4 × wmAPphr. Furthermore, mR@K is reported to exhibit the capability of methods to solve the biased problem of methods.

\textbf{Implementation Details.} 
For all three datasets, we follow the same setup. A deformable DETR with backbone ResNet-50 \cite{he2016deep} is pre-trained for 25 epochs. The number of entity queries, $N_e$, is set to 200. Subsequently, the pre-trained object detector is frozen during the joint training of the predicate decoder and ISD. In the Salience-SGG, the loss ratio between salience loss $\mathcal{L}_{salience}$ and predicate loss $\mathcal{L}_{pre}$ is $1:1$. The re-weighting loss proposed in \cite{wang2021seesaw} is employed as a debiasing strategy. By default, the hyperparameters $\alpha$ and $\beta$ in the loss function are set to 1.0 and 0.2, respectively. The hyper-parameters $\mathcal{T}$ \text{and} $L$ introduced by ISD are 0.6 and 4. The training epochs of predicate decoder and ISD are 13, 8, 12 for VG, OIv6, GQA-200, respectively. We use AdamW \cite{loshchilov2017decoupled} as optimizer, with an initial learning rate of $10^{-4}$ and a weight decay of $10^{-4}$. All training is performed on two NVIDIA RTX A6000 GPUs with a batch size of 16, using a fixed random seed of 42. Additional training details are provided in the \textbf{suppl. material}.
\begin{table*}[t]  
  \centering
  \resizebox{\textwidth}{!}{
  \begin{tabular}{@{} l|c c|c c c|c c c|c c c @{}}
    \toprule
    Method & Backbone & \# params (M)
    & R@20 & R@50 & R@100
    & mR@20 & mR@50 & mR@100
    & F@20 & F@50 & F@100 \\
    \midrule
    \multicolumn{12}{c}{\textit{two-stage models}} \\
    Motifs$^{*}$ \cite{zellers2018neural} {\textcolor{gray}{[CVPR2018]}}
      & ResNeXt101-FPN & 369.9
      & \textcolor{gray}{25.5} & \textcolor{gray}{32.8} & \textcolor{gray}{37.2}
      & 5.0 & 6.8 & 7.9
      & 8.4 & 11.3 & 13.0 \\
    TDE$^{\dag}$ \cite{tang2020unbiased} {\textcolor{gray}{[CVPR2020]}}
      & ResNeXt101-FPN & 369.9
      & \textcolor{gray}{11.9} & \textcolor{gray}{16.6} & \textcolor{gray}{20.2}
      & 6.6 & 8.9 & 11.0
      & 8.5 & 11.7 & 14.3 \\
    BGNN \cite{li2021bipartite} {\textcolor{gray}{[CVPR2021]}}
      & ResNeXt101-FPN & 341.9
      & \textcolor{gray}{23.3} & \textcolor{gray}{31.0} & \textcolor{gray}{35.8}
      & 7.5 & 10.7 & 12.6
      & 11.3 & 15.9 & 18.6 \\
    SHA \cite{dong2022stacked} {\textcolor{gray}{[CVPR2022]}}
      & ResNeXt101-FPN & --
      & \textcolor{gray}{--} & \textcolor{gray}{14.9} & \textcolor{gray}{18.2}
      & \textbf{14.2} & \underline{17.9} & \underline{20.9}
      & -- & 16.3 & 19.5 \\
    IETrans$^{\dag}$ \cite{zhang2022fine} {\textcolor{gray}{[ECCV2022]}}
      & ResNeXt101-FPN & 369.9
      & \textcolor{gray}{17.5} & \textcolor{gray}{23.5} & \textcolor{gray}{27.3}
      & 11.0 & 15.7 & 18.2
      & \textbf{13.5} & \underline{18.8} & \underline{21.8} \\
    ST-SGG$^{\dag}$ \cite{kim2024adaptive} {\textcolor{gray}{[ICLR2024]}}
      & ResNeXt101-FPN & --
      & \textcolor{gray}{--} & \textcolor{gray}{26.7} & \textcolor{gray}{30.7}
      & -- & 11.6 & 14.2
      & -- & 16.2 & 19.4 \\
    DRM \cite{li2024leveraging} {\textcolor{gray}{[CVPR2024]}}
      & ResNeXt101-FPN & --
      & \textcolor{gray}{--} & \textcolor{gray}{19.0} & \textcolor{gray}{22.9}
      & -- & \textbf{20.4} & \textbf{24.1}
      & -- & \textbf{20.8} & \textbf{23.5} \\
    SRD$^{\dag}$ \cite{nguyen2025effective} {\textcolor{gray}{[WACV2025]}}
      & ResNeXt101-FPN & --
      & \textcolor{gray}{--} & \textcolor{gray}{--} & \textcolor{gray}{--}
      & \underline{13.5} & 17.9 & 20.6
      & -- & -- & -- \\
    \midrule
    \multicolumn{12}{c}{\textit{one-stage models}} \\
    SSR-CNN \cite{teng2022structured} {\textcolor{gray}{[CVPR2022]}}
      & ResNeXt101-FPN & 274.3
      & \textcolor{gray}{18.4} & \textcolor{gray}{23.3} & \textcolor{gray}{26.5}
      & \textbf{13.5} & \underline{17.9} & \underline{21.4}
      & \underline{15.6} & 20.2 & 23.7 \\
    EGTR \cite{im2024egtr} {\textcolor{gray}{[CVPR2024]}}
      & ResNet50 & 42.5
      & \textcolor{gray}{22.4} & \textcolor{gray}{28.2} & \textcolor{gray}{31.7}
      & 8.8 & 14.0 & 18.3
      & 12.6 & 18.7 & 23.2 \\
    Mg-RMPN \cite{wang2024multi} {\textcolor{gray}{[ECCV2024]}}
      & ResNet50 & --
      & \textcolor{gray}{22.5} & \textcolor{gray}{29.1} & \textcolor{gray}{33.5}
      & 10.3 & 14.4 & 17.3
      & 14.1 & 19.3 & 22.8 \\
    SpeaQ \cite{kim2024groupwise} {\textcolor{gray}{[CVPR2024]}}
      & ResNet101 & 93.4
      & \textcolor{gray}{25.1} & \textcolor{gray}{32.1} & \textcolor{gray}{35.5}
      & 10.1 & 15.1 & 17.6
      & 14.4 & 20.5 & 23.5 \\
    Hydra-SGG \cite{chen2024hydra} {\textcolor{gray}{[ICLR2025]}}
      & ResNet50 & 67.6
      & \textcolor{gray}{21.9} & \textcolor{gray}{28.6} & \textcolor{gray}{33.4}
      & 10.3 & 15.9 & 19.4
      & 14.0 & \underline{20.5} & \underline{24.7} \\
    \textbf{Salience-SGG (Ours)}
      & ResNet50 & 77.7
      & \textcolor{gray}{21.9} & \textcolor{gray}{28.8} & \textcolor{gray}{33.4}
      & \underline{12.8} & \textbf{18.0} & \textbf{21.6}
      & \textbf{16.2} & \textbf{22.1} & \textbf{26.2} \\
    \bottomrule
  \end{tabular}
  }
  \caption{Comparison with state-of-the-art methods evaluated on the VG test dataset. The methods are divided into two groups. The best and second-best results in each group are indicated with \textbf{bold} and \underline{underlined} text, respectively. `*' denotes the performance without any debiasing strategy. `$^\dag$` indicates the methods are combined with MOTIFS \cite{zellers2018neural}.}
  \label{tab:vg-test}
\end{table*}

\subsection{Comparison to SOTA Methods}
 We compare our approach with both two-stage methods (\eg Motifs \cite{zellers2018neural}, BGNN \cite{li2021bipartite}, DRM \cite{li2024leveraging}, SHA \cite{dong2022stacked}) and one-stage methods (\eg Mg-RMPN \cite{wang2024multi}, SSR-CNN \cite{teng2022structured}, SpeaQ \cite{kim2024groupwise}, Hydra-SGG \cite{chen2024hydra}). Since our works are based on Unbiased-SGG methods. we focus on comparing to results in which at least one debiasing strategy is applied.

\begin{table}[t]
  \centering
  \resizebox{\linewidth}{!}{
  \begin{tabular}{@{} c|c c |c c |c @{}}
    \toprule
    Method & mR@50 & micro-R@50 & wmAPrel & wmAPphr & score \\
    \midrule
    \multicolumn{6}{c}{\textit{two-stage models}} \\
    GPS-Net \cite{lin2020gps} {\textcolor{gray}{[CVPR2020]}} & 35.2 & \textcolor{gray}{74.8} & 32.9 & 34.0 & 41.7 \\
    BGNN \cite{li2021bipartite} {\textcolor{gray}{[CVPR2021]}} & 40.5 & \textcolor{gray}{75.0} & 33.5 & 34.2 & 42.1 \\
    RU-Net \cite{lin2022ru} {\textcolor{gray}{[CVPR2022]}} & -- & \textcolor{gray}{76.9} & 35.4 & 34.9 & 43.5 \\
    PE-Net \cite{zheng2023prototype} {\textcolor{gray}{[CVPR2023]}} & -- & \textcolor{gray}{76.5} & \underline{36.6} & \underline{37.4} & \underline{44.9} \\
    DRM \cite{li2024leveraging} {\textcolor{gray}{[CVPR2024]}} & -- & \textcolor{gray}{75.9} & \textbf{40.5} & \textbf{41.4} & \textbf{47.9} \\
    \midrule
    \multicolumn{6}{c}{\textit{one-stage models}} \\
    SGTR \cite{Li_2022_CVPR} {\textcolor{gray}{[CVPR2022]}} & 42.6 & \textcolor{gray}{59.9} & 37.0 & 38.7 & 42.3 \\
    SSR-CNN \cite{teng2022structured} {\textcolor{gray}{[CVPR2022]}} & 42.8 & \textcolor{gray}{76.7} & 41.5 & 43.6 & 49.4 \\
    RelTR \cite{cong2023reltr} {\textcolor{gray}{[TPAMI2023]}} & -- & \textcolor{gray}{71.7} & 37.2 & 37.5 & 43.0 \\
    EGTR \cite{im2024egtr} {\textcolor{gray}{[CVPR2024]}} & -- & \textcolor{gray}{75.0} & 42.0 & 41.9 & 48.6 \\
    Mg-RMPN \cite{wang2024multi} {\textcolor{gray}{[ECCV2024]}} & \underline{45.5} & \textcolor{gray}{74.2} & 35.5 & 36.4 & 43.6 \\
    Hydra-SGG \cite{chen2024hydra} {\textcolor{gray}{[ICLR2025]}} & -- & \textcolor{gray}{76.1} & \underline{42.8} & \underline{44.3} & \underline{50.1} \\
    \textbf{Salience-SGG (Ours)} & \textbf{48.0} & \textcolor{gray}{78.1} & \textbf{45.6} & \textbf{44.9} & \textbf{51.8} \\
    \bottomrule
  \end{tabular}
  }
  \caption{Comparison with SOTA methods evaluated on OIv6.}
  \label{tab:OIv6}
\end{table}

\begin{table}[t]
  \centering
  \resizebox{\linewidth}{!}{
  \begin{tabular}{@{} c|c c |c c |c c @{}}
    \toprule
    Method & R@50 & R@100 & mR@50 & mR@100 & F@50 & F@100 \\
    \midrule
    \multicolumn{7}{c}{\textit{two-stage models}} \\
    Motifs$^{\dag}$ \cite{zellers2018neural} {\textcolor{gray}{[CVPR2018]}} 
    & \textcolor{gray}{18.5} & \textcolor{gray}{21.8} & 16.8 & 18.8 & \underline{17.6} & \underline{20.2} \\
    VCTREE$^{\dag}$ \cite{tang2019learning} {\textcolor{gray}{[CVPR2019]}} 
    & \textcolor{gray}{17.6} & \textcolor{gray}{20.7} & 15.6 & 17.8 & 16.5 & 19.1 \\
    SHA$^{\dag}$ \cite{dong2022stacked} {\textcolor{gray}{[CVPR2022]}} 
    & \textcolor{gray}{14.8} & \textcolor{gray}{17.9} & \underline{17.8} & \underline{20.1} & 16.2 & 18.9 \\
    VETO \cite{sudhakaran2023vision} {\textcolor{gray}{[ICCV2023]}} 
    & \textcolor{gray}{26.1} & \textcolor{gray}{29.0} & 7.0 & 8.1 & 11.0 & 12.7 \\
    DRM \cite{li2024leveraging} {\textcolor{gray}{[CVPR2024]}} 
    & \textcolor{gray}{18.6} & \textcolor{gray}{21.7} & \textbf{18.9} & \textbf{21.0} & \textbf{18.7} & \textbf{21.3} \\
    RA-SGG \cite{yoon2025ra} {\textcolor{gray}{[AAAI2025]}} 
    & \textcolor{gray}{16.3} & \textcolor{gray}{19.0} & 12.9 & 15.0 & 14.4 & 16.8 \\
    \midrule
    \multicolumn{7}{c}{\textit{one-stage models}} \\
    Pair-Net \cite{wang2024pair} {\textcolor{gray}{[PAMI2024]}} 
    & \textcolor{gray}{20.2} & \textcolor{gray}{23.4} & 10.6 & 12.6 & 13.9 & 16.4 \\
    Mg-RMPN \cite{wang2024multi} {\textcolor{gray}{[ECCV2024]}} 
    & \textcolor{gray}{23.2} & \textcolor{gray}{25.7} & \underline{12.8} & 14.5 & \underline{16.5} & 18.5 \\
    Hydra-SGG \cite{chen2024hydra} {\textcolor{gray}{[ICLR2025]}} 
    & \textcolor{gray}{22.8} & \textcolor{gray}{26.5} & 12.7 & \underline{15.9} & 16.3 & \underline{19.9} \\
    \textbf{Salience-SGG (Ours)} 
    & \textcolor{gray}{23.6} & \textcolor{gray}{26.6} & \textbf{16.2} & \textbf{18.4} & \textbf{19.2} & \textbf{21.7} \\
    \bottomrule
  \end{tabular}
  }
  \caption{Comparison with SOTA methods evaluated on GQA-200. `$^\dag$` denotes models combined with GCL as proposed in \cite{dong2022stacked}.}
  \label{tab:GQA-200}
\end{table}

\textbf{VG.} In \Cref{tab:vg-test}, we report the comparison results on the VG test dataset. The Salience-SGG model demonstrates superior holistic performance F@K compared to all existing SGG models which incorporate debiasing strategies. In the context of one-stage models, the Salience-SGG model achieves second best mR@20 with 12.8 and the SOTA mR@50, and mR@100 of 18.0 and 21.6, respectively. This is notable given that our model utilizes a significantly reduced number of parameters (77.7M) compared to the SSR-CNN model \cite{teng2022structured} (274.3M). In contrast to the Mg-RMPN approach \cite{wang2024multi}, which also employs seesaw loss as the debiasing strategy, our approach significantly improves mR@K. This demonstrates that our salience estimation process enhances addressing the long-tailed distribution problem. Among two-stage models, DRM \cite{li2024leveraging} exhibits notable mR@50 and mR@100 scores of 20.4 and 24.1, respectively. However, weak robustness to the debiasing strategy introduces substantial negative impacts on R@K by preventing it from achieving a superior overall performance F@K. We provide specific comparisons to evaluate the robustness (see \Cref{Complementary Comparisons}). In addition, a more extensive comparison with additional methods is provided in \textbf{suppl. material}.

\textbf{OIv6.} For the OIv6 dateset, a different set of evaluation metrics is applied. Since micro-R@50, wmAPrel, wmAPphr are more sensitive to the performance on frequent predicates, most existing works, except Mg-RMPN \cite{wang2024multi}, omits debiasing strategies to evaluate their models, resulting in better results. In Salience-SGG, the debiasing strategy is maintained to illustrate the robustness of our framework to the debiasing strategies. In addition, we follow \cite{Li_2022_CVPR} and report mR@K to reflect the performance on rare predicates. As demonstrated in \Cref{tab:OIv6}, our approach sets new SOTA standards on all evaluation metrics. It is imperative to emphasize the result on wmAPrel, which is employed to assess the AP of the predicted triplets, wherein both the subject and object boxes exhibit an IoU of at least 0.5 with the ground-truth. Our approach outperforms the second best result from Hydra-SGG \cite{chen2024hydra} by a margin of 2.8 on wmAPrel. This may stem from the relative geometric relationships encoded in the triplet salience labels.

\textbf{GQA-200.} \Cref{tab:GQA-200} illustrates the comparison results on the GQA-200 test dataset. As with the VG dataset, the Salience-SGG sets new SOTA F@k scores of 19.2 and 21.7, respectively. This outcome surpasses the second best results from SHA \cite{dong2022stacked} and DRM \cite{li2024leveraging}. Furthermore, Salience-SGG outperforms one-stage models Mg-RMPN \cite{wang2024multi} and Hydra-SGG \cite{chen2024hydra} on all evaluation metrics.

\subsection{Complementary Comparisons}
\label{Complementary Comparisons}
This section presents additional evaluations on the VG test set to assess (1) whether Salience-SGG exhibits superior robustness to debiasing compared to Unbiased-SGG models, (2) the compatibility of our salience estimation framework with alternative debiasing strategies, and (3) the performances of our ISD on predicates at varying frequencies.

\textbf{Robustness Evaluation.} 
Debiasing strategies can shift the trade-off between mR@K and R@K. When two approaches differ greatly in mR@K, F@K alone may not reflect their robustness.
To assess robustness, we train Salience-SGG on VG with varying seesaw loss hyperparameters, yielding results across different mR@K values.
Since mR@K is more sensitive to $\beta$, we vary $\beta$ while keeping $\alpha$ fixed. For a fair comparison, we select the Salience-SGG result with the closest mR@K, so differences in F@K more clearly reflect robustness to debiasing.
\Cref{tab:Robustness} reports Salience-SGG with different $\beta$ values and three strong baselines. Comparing Salience-SGG $\beta=0.35$, $0.25$, and $0.125$ with DRM \cite{li2024leveraging}, SHA \cite{dong2022stacked}, and SpeaQ \cite{kim2024groupwise}, respectively, highlights Salience-SGG’s robustness to debiasing.
Benefiting from our salience estimation process, our Salience-SGG set new SOTA mR@50 and mR@100 scores with 21.4 and 25.3 in the case where $\beta = 0.5$.

\begin{figure}[t]
    \centering
    \includegraphics[width=\linewidth]{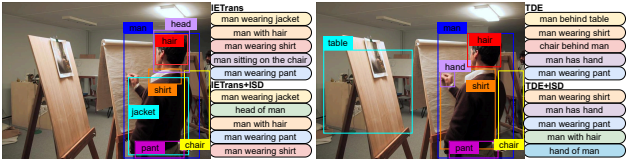}
    \caption{Comparison of the top-5 detected triplets. For each image, the colors indicate triplet identifications; identical colors before and after re-ranking refer to the same triplets.
    }
    \label{fig: visulization}
\end{figure}

\textbf{ISD Compatibility with Other Debiasing Strategies.}
To examine the compatibility of triplet salience estimation with diverse debiasing strategies, the ISD is integrated into the pre-trained TDE \cite{tang2020unbiased} and IETrans \cite{zhang2022fine}. The original detection set is re-ranked by involving ISD salience scores. \Cref{tab:other debiasing} shows that our salience estimation improves both TDE and IETrans, demonstrating the ISD module’s broad compatibility with debiasing strategies. We also provide the qualitative comparisons for TDE versus TDE+ISD, and IETrans versus IETrans+ISD. In order to provide a comprehensive understanding, we visualize the top-5 detections from each models as illustrated in \Cref{fig: visulization}. For both models, the spatial incoherent triplets such as $\{man, sitting\_on, chair\}$, $\{man, behind, table\}$ are removed after re-ranking. The triplet $\{man, with, hair\}$ detected by IETrans and IETrans+ISD shows that re-ranking is not simply the result of lowering confidences for rare predicates. For more qualitative comparisons please see \textbf{suppl. material}.

\begin{table}  
  \centering
  \resizebox{\linewidth}{!}{
  \begin{tabular}{
    @{} c |c c |c c |c c
    @{}
  }
    \toprule
    Method&  {R@50} & {R@100} &{mR@50} & {mR@100} & {F@50}&{F@100}\\
    \midrule
    DRM \cite{li2024leveraging} & 19.0 & 22.9 & 20.4 & 24.1 & 20.8 & 23.5 \\
    SHA \cite{dong2022stacked}& 14.9 & 18.2 & 17.9 & 20.9& 16.3 & 19.5 \\
    SpeaQ \cite{kim2024groupwise} & 32.1 & 35.5 & 15.1 & 17.6 & 20.5 & 23.5 \\
    Salience-SGG ($\beta=0.5$) & 20.3 & 24.3 & 21.4& 25.3 &20.8& 24.8 \\
    Salience-SGG ($\beta=0.35$) & 24.0 & 28.4 & 20.4& 24.4 &22.0& 26.2 \\
    Salience-SGG ($\beta=0.25$)& 26.8 & 31.4 & 19.2& 22.9 &22.4& 26.5 \\
    Salience-SGG ($\beta=0.125$) & 32.2 & 36.2 & 15.7& 19.0 &21.1& 24.9 \\
    \bottomrule
  \end{tabular}
  }
  \caption{Robustness to debiasing strategies evaluation.}
  \label{tab:Robustness}
\end{table}

\begin{table}  
  \centering
  \resizebox{\linewidth}{!}{
  \begin{tabular}{
    @{} c |c c |c c |c c
    @{}
  }
    \toprule
    Method&  {R@50} & {R@100} &{mR@50} & {mR@100} & {F@50}&{F@100}\\
    \midrule
    TDE \cite{tang2020unbiased} & 16.6 & 20.2 & 8.9 & 11.0& 11.7 & 14.3 \\
    TDE+\textbf{ISD} & 19.9 & 24.4 & 10.5 & 12.4& 13.6 & 16.4 \\
    IETrans \cite{zhang2022fine} & 23.5 & 27.3 & 15.7 & 18.2 & 18.8 & 21.8 \\
    IETrans+\textbf{ISD} & 27.6 & 32.5 & 15.9& 18.6 &20.2& 23.7 \\
    \bottomrule
  \end{tabular}
  }
  \caption{Results of combing ISD with existing two-stage Unbiased-SGG models.}
  \label{tab:other debiasing}
\end{table}

\begin{table}  
  \centering
  \resizebox{\linewidth}{!}{
  \begin{tabular}{
    @{} c |c c c c
    @{}
  }
    \toprule
    Model&  {Head (16)} & {Body (17)} &{Tail (17)} &{mR@100}\\
    \midrule
    DT2-ACBS \cite{desai2021learning}&22.3&26.7&\bf{24.0}  &24.4\\
    EGTR \cite{im2024egtr} & 24.3&19.4&13.3 &18.9\\
    TDE \cite{tang2020unbiased} & 19.8 & 13.1 & 0.1 & 11.0\\
    IETrans \cite{zhang2022fine}&23.8&18.0&13.2&18.2\\
    ST-SGG \cite{kim2024adaptive} & 21.5 & 15.0& 0.1& 14.2\\
    BGNN \cite{li2021bipartite}& 22.3&11.9&0.1&12.5\\
    \midrule
    TDE+ISD & 21.6 &14.3 &0.1&12.4\\
    IETrans+ISD & 26.1 &19.2 &11.0 &18.6\\
    Salience-SGG ($\beta=0.5$) & 25.9 & \bf{28.0} & 22.1 &\bf{25.3} \\
    Salience-SGG($\beta=0.2$) & \bf{27.4} & 22.8 & 15.0 &21.6 \\
    \bottomrule
  \end{tabular}
  }
  \caption{Group-wise performance comparison}
  \label{Tab: Group-wise performance comparison}
\end{table}

\textbf{Group-wise Performance Comparison.} 
Following \cite{desai2021learning}, predicates are grouped into head, body, and tail based on training frequency, and the performance of each group is measured by the average mR@100.
\Cref{Tab: Group-wise performance comparison} shows that integrating ISD significantly improves TDE and IETrans on the head and body groups. Although IETrans+ISD scores are lower on the tail group, overall performance is still improved. One-stage Salience-SGG achieves SOTA on head and body groups with superior overall performance.


\subsection{Salience Sensitivity Analysis}
\label{Spatial Understanding Analysis}

\begin{figure}[t] 
  \centering
    \begin{tikzpicture}
      \begin{axis}[
          width=\linewidth,
          height=0.65\linewidth,
          xlabel={\textbf{F@100}},
          ylabel={\textbf{pl-AP}},
          grid=major,
          font=\small,
          xmin=10, xmax=30,
          ymin=10, ymax=35,
          legend style={at={(1,0)}, anchor=south east, draw=black, font=\tiny},
          every axis y label/.append style={yshift=-10pt}
      ]
        
        
        \addplot[mark=square*,draw=black,fill=none,mark size=3pt] coordinates {(0,0)};
        \addlegendentry{SGG Models}
        
        \addplot[mark=diamond*,draw=black,fill=none,mark size=3pt] coordinates {(0,0)};
        \addlegendentry{Unbiased-SGG Models}
        
        \addplot[mark=diamond*,draw=blue,fill=blue,mark size=5pt] coordinates {(14.3,11.5)} node[right,font=\scriptsize, xshift=1pt] {TDE \cite{tang2020unbiased}};
        \addplot[mark=diamond*,draw=purple,fill=purple,mark size=5pt] coordinates {(18.6,22.0)} node[right,font=\scriptsize,xshift=1pt] {BGNN \cite{li2021bipartite}};
        \addplot[mark=diamond*,draw=blue,fill=blue,mark size=5pt] coordinates {(19.4,18.6)} node[right,font=\scriptsize, xshift=1pt] {ST-SGG \cite{kim2024adaptive}};
        \addplot[mark=diamond*,draw=blue,fill=blue,mark size=5pt] coordinates {(21.8,25.1)} node[right,font=\scriptsize, xshift=1pt] {IETrans \cite{zhang2022fine}};
        \addplot[mark=diamond*,draw=blue,fill=blue,mark size=5pt] coordinates {(23.7,27.4)} node[right,font=\scriptsize, xshift=1pt] {IETrans+\textcolor{green}{\textbf{ISD}}};
        \addplot[mark=square*,draw=blue,fill=blue,mark size=4pt] coordinates {(13.0,25.1)} node[right,font=\scriptsize, xshift=2pt] {Motifs \cite{zellers2018neural}};
        \addplot[mark=diamond*,draw=green,fill=green,mark size=5pt] coordinates {(26.2,30.8)} node[above,font=\scriptsize,yshift=2pt,color=green] {\textbf{Salience-SGG}};
        \addplot[mark=diamond*,draw=orange,fill=orange,mark size=5pt] coordinates {(23.2,28.1)} node[above,font=\scriptsize, yshift=2pt] {EGTR \cite{im2024egtr}};
        \addplot[mark=square*,draw=orange,fill=orange,mark size=4pt] coordinates {(15.6,29.4)} node[above,font=\scriptsize, yshift=2pt] {EGTR \cite{im2024egtr}};
      \end{axis}
    \end{tikzpicture}
    \caption{Pairwise localization avg. precision (pl-AP) versus SGG performance (F@100). Color indicates the model family. The green text marks our contributions}
    \label{fig: spatial understanding analysis}
\end{figure}
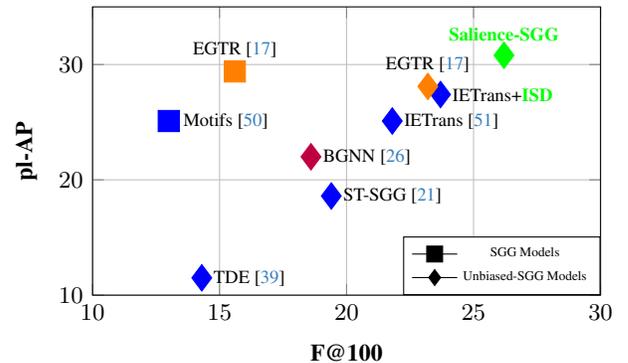

Pairwise Localization Average Precision (pl-AP) measures the model’s ability to localize salient subject–object pairs in predicted triplets.
For the top-100 detected triplets, we compute category-agnostic precision and recall, where precision reflects accuracy in capturing salient spatial structure and recall reflects its coverage. Similar to the saliency triplet label assignment, a true positive is assigned if the IoU between subject and object pairs is at least 0.5. The pl-AP is then calculated as the area under the precision-recall curve.
We report pl-AP with the SGG performance (F@100) of multiple SGG and Unbiased-SGG models, as illustrated in \Cref{fig: spatial understanding analysis}. From this figure, the following observations can be made: 1) The comparisons of pl-AP and F@100 among the Unbiased-SGG models clearly demonstrate that a higher pl-AP leads to a better F@100. 2) The debiasing strategies tend to impede the capture of salient spatial structures. In particular, Motifs \cite{zellers2018neural} achieves a higher pl-AP than TDE \cite{tang2020unbiased} and ST-SGG \cite{kim2024adaptive}. Of note, IETrans \cite{zhang2022fine} achieves a similar pl-AP to Motifs despite IETrans introducing spatial incoherent triplets (\ie, lower precision). This is likely the result of increased diversity of captured salient structures (\ie, higher recall). 3) Comparing IETrans and IETrans+ISD highlights that ISD not only improves F@100 but also leads to an increased pl-AP.
\subsection{Ablation Studies}
\label{Ablation Study}
Finally, in this section, we present ablation studies to provide detailed analyses of our approach. All experiments are conducted on the VG dataset. The analysis of hyperparameters (\ie $\mathcal{T}$ \text{and} $L$) is provided in \textbf{suppl. material}.

\textbf{Impact of Individual Components.} We propose an ISD module containing G-ESA and P-ECA. \Cref{tab:proposed components} illustrates the performance when each component is removed. When G-ESA and P-ECA are partially removed, we use the regular self-attention and cross-attention layers instead to maintain the module at the same parameter level. If the ISD is completely removed, performance drops significantly. This demonstrates that our ISD can greatly improve the performance of Unbiased-SGG. Partial removal of the G-ESA and P-ECA modules causes a decline in both R@K and mR@K. These results highlight the importance of our G-ESA and P-ECA modules. The G-ESA may facilitate spatial structure learning, while the P-ECA enhances predicate understanding (\ie the salience loss serves as a regularization for the predicate prediction through the P-ECA). A detailed analysis of predicate understanding and spatial structure learning is provided in the \textbf{suppl. material}.

\textbf{Iterative Refinement Analysis.} \Cref{tab:iterative refinement} demonstrates that an iterative refinement process in ISD leads to a better performance than the non-iterative process.

 \begin{table}  
  \centering
  \resizebox{\linewidth}{!}{
  \begin{tabular}{
    @{} c c c |c c |c c |c c
    @{}
  }
    \toprule
    ISD& G-ESA& P-ECA&  {R@50} & {R@100} &{mR@50} & {mR@100} & {F@50}&{F@100}\\
    \midrule
    $\times$ & $\times$ & $\times$ & 26.4 & 29.7 & 13.5 & 16.4& 17.8 & 21.1\\
    $\checkmark$ & $\checkmark$ & $\times$ & 27.0 & 31.2 & 17.8 & 20.6& 21.5 & 24.9 \\
    $\checkmark$ & $\times$ & $\checkmark$ & 27.8 & 32.3 & 17.1 & 20.5& 21.2 & 25.1 \\
    $\checkmark$ & $\checkmark$ & $\checkmark$ & 28.8 & 33.4 & 18.0 & 21.6& 22.1 & 26.2 \\
    \bottomrule
  \end{tabular}
  }
  \caption{Ablation study on the individual model components.}
  \label{tab:proposed components}
\end{table}

\begin{table}  
  \centering
  \resizebox{\linewidth}{!}{
  \begin{tabular}{
    @{} c |c c |c c |c c
    @{}
  }
    \toprule
    Iterative Refinement&  {R@50} & {R@100} &{mR@50} & {mR@100} & {F@50}&{F@100}\\
    \midrule
    $\times$ & 28.6 & 33.0 & 17.1 & 20.7& 21.5 & 25.4 \\
    $\checkmark$ & 28.8 & 33.4 & 18.0 & 21.6& 22.1 & 26.2 \\
    \bottomrule
  \end{tabular}
  }
  \caption{Ablation study on the iterative refinement process in ISD.}
  \label{tab:iterative refinement}
\end{table}

\begin{table}  
  \centering
  \resizebox{\linewidth}{!}{
  \begin{tabular}{
    @{} c |c c |c c |c c
    @{}
  }
    \toprule
    Type&  {R@50} & {R@100} &{mR@50} & {mR@100} & {F@50}&{F@100}\\
    \midrule
    $\text{top\_down}_{\text{gt}}$ & 27.5 & 32.0&15.1&18.4&19.5&23.3
     \\
     $\text{top\_down}_{\text{entity}}$ &28.5& 33.0& 16.3 & 19.6&20.7& 24.6\\
     $\text{top\_down}_{\text{triplet}}$ &27.9& 32.5& 15.7 & 19.1&20.1& 24.1\\
     Ours & 28.8 & 33.4 & 18.0 & 21.6& 22.1 & 26.2\\
    \bottomrule
  \end{tabular}
  }
  \caption{Ablation study on the type of salience labels.}
  \label{tab:type of salience label}
\end{table}

\textbf{Triplet Salience Label Analysis.} To investigate the effect of our bottom-up triplet salience labels, we create three types of top-down labels which consider the semantic information concurrently, namely ${top\_down}_{gt}$, ${top\_down}_{entity}$ and ${top\_down}_{triplet}$. ${top\_down}_{gt}$ is created by mapping the predicate label $G$ to a binary mask as in previous work \cite{im2024egtr, wang2024multi, jung2023devil}. Following \cite{chen2024hydra}, ${top\_down}_{entity}$ is created by performing an additional one-to-many matching on the object detector results. ${top\_down}_{triplet}$ is constructed by performing one-to-many matching at the triplet level as in \cite{kim2024groupwise}. For all triplet salience labels, the predicate label $G^{'}$ is consistently obtained by the one-to-one matching results from the object detector. The details of the construction process of each type of triplet salience label are specified in the \textbf{suppl. material}. \Cref{tab:type of salience label} shows that Salience-SGG, which is supervised by bottom-up triplet salience labels offering maximal spatial supervision, achieves the best results.

%% file: sec/5_conclusion.tex
\section{Conclusion}
\label{Conclusion}
The present study provides a detailed examination of the underlying causes of the substantial performance deterioration of Unbiased-SGG models on frequent predicates. Based on our observations, we propose a framework incorporating an iterative triplet salience estimation process to enhance the Unbiased-SGG process. The salience estimation process, supervised by semantic-agnostic salience labels, enables Unbiased-SGG models to globally highlight the triplets characterized by salient spatial structures. This improvement is demonstrated quantitatively by a specific evaluation metric that measures a model's ability to capture salient triplet structures. To model the process of salience estimation, two enhanced attention layers are devised to direct the salience message-passing. Extensive experimentation and ablation studies are conducted, demonstrating that the introduced framework achieves SOTA performance and exhibits robustness to diverse debiasing strategies. \\

\section*{Acknowledgement}
Funded by the Deutsche Forschungsgemeinschaft (DFG, German Research Foundation) under Germany’s Excellence Strategy – EXC 2002/1 “Science of Intelligence” – project number 390523135. In addition, this work was partially supported by MIAI@Grenoble Alpes, (ANR-19-P3IA-0003).